\documentclass[sigconf]{acmart}

\usepackage{algorithm}
\usepackage{multirow}
\usepackage{appendix}
\usepackage{enumitem} 
\usepackage{subcaption}
\usepackage{algorithmic}
\DeclareMathOperator*{\argmax}{argmax}

\newcommand{\method}{\texorpdfstring{{\color{black}Gaze-guided Action Anticipation}}{Gaze-guided Action Anticipation}}

\AtBeginDocument{%
  }
\title{Gaze-Guided Graph Neural Network for Action Anticipation Conditioned on Intention}

\copyrightyear{2024}
\acmYear{2024}
\setcopyright{acmlicensed}\acmConference[ETRA '24]{2024 Symposium on Eye Tracking Research and Applications}{June 4--7, 2024}{Glasgow, United Kingdom}
\acmBooktitle{2024 Symposium on Eye Tracking Research and Applications (ETRA '24), June 4--7, 2024, Glasgow, United Kingdom}
\acmDOI{10.1145/3649902.3653340}
\acmISBN{979-8-4007-0607-3/24/06}

\begin{document}

\author{Suleyman Ozdel}
\authornote{Both authors contributed equally to the paper.}
\affiliation{%
  \institution{Technical University of Munich}
  \city{Munich}
  \country{Germany}
}
\email{ozdelsuleyman@tum.de}
\orcid{0000-0002-3390-6154}

\author{Yao Rong}
\authornotemark[1]
\affiliation{%
  \institution{Technical University of Munich}
  \city{Munich}
  \country{Germany}
}
\email{yao.rong@tum.de}
\orcid{0000-0002-6031-3741}

\author{Berat Mert Albaba}
\affiliation{%
  \institution{ETH}
  \city{Zurich}
  \country{Switzerland}
}
\email{mert.albaba@inf.ethz.ch}
\orcid{0000-0002-3406-8412}

\author{Yen-Ling Kuo}
\affiliation{%
  \institution{University of Virginia}
  \city{Charlottesville}
  \country{United States}
}
\email{ylkuo@virginia.edu}
\orcid{0000-0002-6433-6713}

\author{Xi Wang}
\affiliation{%
  \institution{ETH}
  \city{Zurich}
  \country{Switzerland}
}
\email{xi.wang@inf.ethz.ch}
\orcid{0000-0001-5442-1116}

\author{Enkelejda Kasneci}
\affiliation{%
  \institution{Technical University of Munich}
  \city{Munich}
  \country{Germany}}
\email{enkelejda.kasneci@tum.de}
\orcid{0000-0003-3146-4484}

\begin{abstract}
Humans utilize their gaze to concentrate on essential information while perceiving and interpreting intentions in videos. Incorporating human gaze into computational algorithms can significantly enhance model performance in video understanding tasks. In this work, we address a challenging and innovative task in video understanding: predicting the actions of an agent in a video based on a partial video. We introduce the Gaze-guided Action Anticipation algorithm, which establishes a visual-semantic graph from the video input. Our method utilizes a Graph Neural Network to recognize the agent's intention and predict the action sequence to fulfill this intention. To assess the efficiency of our approach, we collect a dataset containing household activities generated in the VirtualHome environment, accompanied by human gaze data of viewing videos. Our method outperforms state-of-the-art techniques, achieving a 7\% improvement in accuracy for 18-class intention recognition. This highlights the efficiency of our method in learning important features from human gaze data.
\end{abstract}

\ccsdesc[300]{Human-centered computing~Human computer interaction (HCI)}
\ccsdesc[500]{Computing methodologies~Activity recognition and understanding}
\ccsdesc[500]{Computing methodologies~Machine learning}

\keywords{Eye-tracking, Human-computer interaction, Action prediction, Action recognition}

\maketitle

\section{Introduction}
\label{sec:intro}
\begin{figure}[ht]
    \centering
    \includegraphics[width=0.95\linewidth]{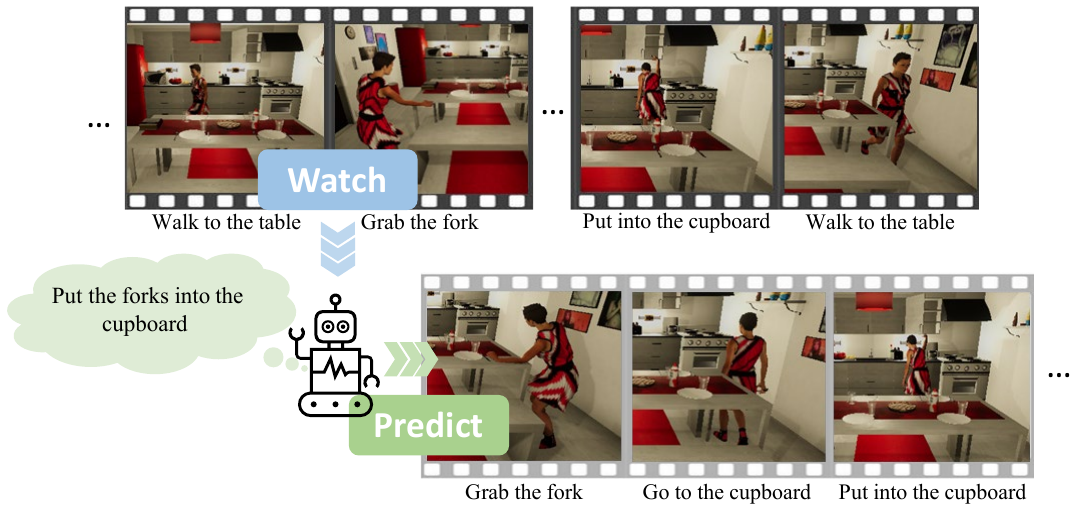}
    \caption{The household activity (intention) of the human agent in the video is to put the cutlery (forks) into the cupboard. In the first \textit{watch} step, the AI agent observes the human activity. In the second \textit{predict} step, the AI agent aims to understand the human intention and predict the remaining actions in order to complete the task.}
    \Description{The household activity (intention) of the human agent in the video is to put the cutlery (forks) into the cupboard. In the first \textit{watch} step, the AI agent observes the human activity. In the second \textit{predict} step, the AI agent aims to understand the human intention and predict the remaining actions in order to complete the task.}
    \label{fig:teaser}
\end{figure}

As AI technology emerges into our daily lives, AI models are designed to assist humans with various tasks in industrial~\cite{villani2018survey}, agricultural~\cite{vasconez2019human}, and household settings~\cite{pham2017evaluating,kraus2022kurt}. For instance, \citet{kraus2022kurt} designs a collaborative system where the robot is able to engage in the task through natural dialogues with the user. In this paper, we tackle a relevant task where an AI agent aims to help humans accomplish household tasks. To do so, the agent first needs to understand the users' intentions. Therefore, we design a prediction task: Given an observation of a user doing certain household activities, predict what the overall activity goal is (human intention) and what the remaining steps are in order to complete this goal. This simulates realistic scenarios of a ``smart home'', where robotic assistants help humans to accomplish tasks without the requirement of specific instructions, as illustrated in Figure \ref{fig:teaser}.

When attempting to understand another person's intentions, humans often rely on contextual information from the environment or objects that the person is interacting with.
Humans use their gaze fixations to precisely extract this information. Previous studies have employed gaze sequences in many tasks, for instance in fine-grained classification~\cite{rong2021human}, computer-aided medical diagnosis systems~\cite{karargyris2021creation}, and object selection/cropping in images~\cite{santella2006gaze}. To understand the intention of another person, it is essential to identify task-relevant  objects and their interconnections, as demonstrated by \citet{panetta2019software}. To equip the model with the ability to perceive, we propose a pipeline to utilize the human attention mechanism (gaze behavior) to  extract essential information and analyze the connections between them.

Previous work~\cite{min2021integrating} proposes to integrate human gaze into an attention module for activity recognition. Their model is applied to ego-centric videos~\cite{li2015delving,li2018eye}, which do not capture the environment well.
In this paper, we integrate human attention into a model designed for activity recognition and anticipation given third-person view videos. 
The primary challenge lies in effectively extracting human fixations to obtain informative insights about the activity, thereby building a foundational understanding. To address this research challenge, we propose a framework, which we have named \method. 
In tackling this challenge, we represent the video using a graph by utilizing human gaze data to extract informative parts of the scenes.
We design an algorithm that uses the fixation points to extract the essential spatial-temporal features. These visual embeddings are turned into nodes. The object detector identifies the objects on which these fixation points fall and the model obtains object features based on their classes (labels). These features are further utilized to build edge features to enhance the graph with interacting object information. 
We train and evaluate our framework using our dataset collected from the household task videos generated from the ``VirtualHome'' simulator~\cite{puig2018virtualhome}. Our results indicate that human attention is indeed beneficial for recognizing activities and predicting the actions to fulfill the goal. The proposed method is able to understand human household activities and anticipate future actions to complete the main goal.
The contributions can be summarized as follows:
\begin{itemize}
    \item  We tackle a challenging task of human action prediction, including human intention recognition and anticipation. We use fixations as guidance to build graphs that encompass task-relevant spatial-temporal features.
    \item Our proposed framework, named \method, utilizes a Graph Neural Network (GNN) to tackle the novel task formulation of predicting atomic actions conditioned on intentions.
    \item We introduce an eye-tracking dataset comprising 185 videos of household activities generated in the environment VirtualHome. These videos contain 178 atomic actions, 18 activity classes, and robust annotations, including semantic segmentation and object interactions, across four different room settings.
    \item Our framework outperforms other state-of-the-art baselines in the human action prediction task, showing the effectiveness of our framework.
\end{itemize}

\section{Related Work}
\label{sec:related work}

Human gaze-based attention has been applied to various AI applications, including autonomous driving~\cite{braunagel2017ready,xia2019predicting,xia2020periphery}, human-robot interaction~\cite{aronson2018eye,shafti2019gaze,weber2020distilling}, fine-grained classification~\cite{rong2021human}, and medical image inspection~\cite{karargyris2021creation}. Several works on activity understanding prove the advantage of utilizing human attention. For instance, \citet{nagarajan2020ego} propose to distill graphs from egocentric videos (EPIC-Kitchens~\cite{damen2018scaling}) and utilizes graphs to predict actions, while \citet{min2021integrating} integrate human attention (gaze fixation) into the model attention mechanism, where the Convolutional Neural Network (CNN)-based attention module is trained by human fixation points. \citet{zheng2022gimo} introduce a new dataset named GIMO, which captures both 3D body poses of the human agent and their eye gaze from an ego-centric view during their daily activities. This dataset facilitates the study of human motion prediction and their result confirm the significance of intent information derived from eye gaze. In this project, we aim to enhance AI models with human attention on videos from a \textit{third-person} view, and we contextualize it via a practical use case involving daily household activities inspired by VirtualHome~\cite{puig2018virtualhome}. 

\begin{figure*}
    \centering
    \includegraphics[width=.95\linewidth]{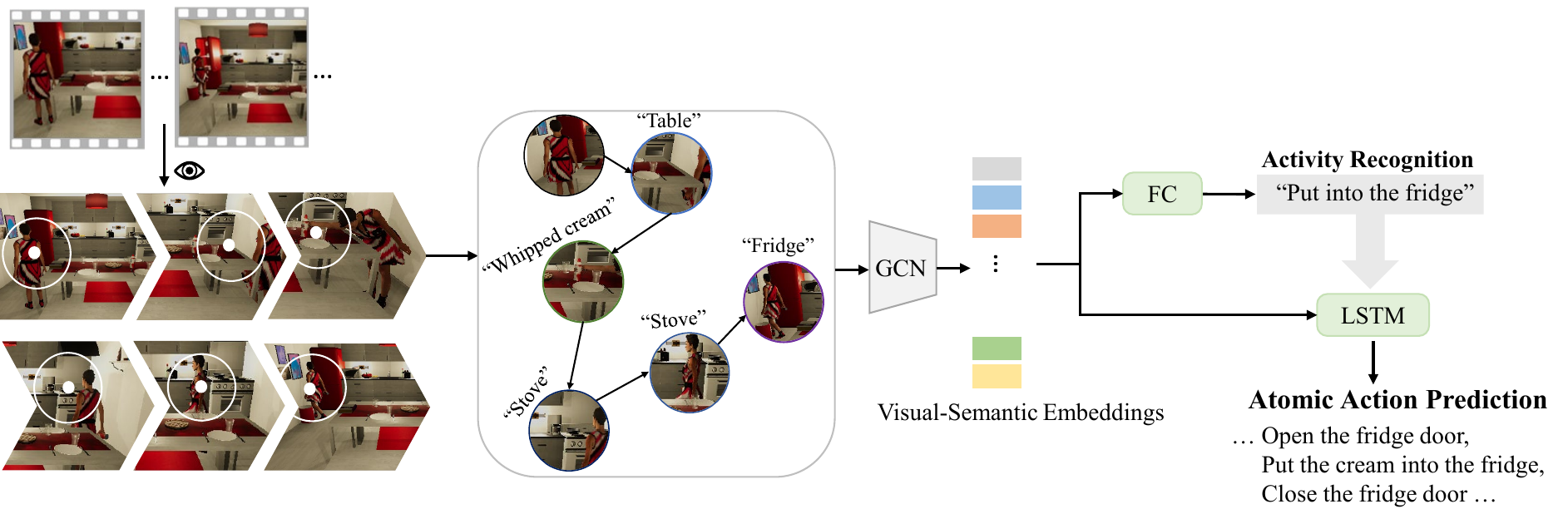}
    \caption{Workflow of our proposed framework, \method. Our model first predicts the gaze fixation (Left) and establishes a visual-semantic graph from the input video (\textbf{Middle}). Based on the graph, it solves the downstream tasks activity recognition and action prediction ({Right}).}
    \Description{Workflow of our proposed framework, \method. Our model first predicts the gaze fixation (Left) and establishes a visual-semantic graph from the input video (\textbf{Middle}). Based on the graph, it solves the downstream tasks activity recognition and action prediction ({Right}).}
    \label{fig:method}
\end{figure*}

\section{VirtualHome Video Dataset}
\label{sec:dataset}
We use VirtualHome~\cite{puig2018virtualhome} to generate videos where a virtual agent performs daily household activities in a rich environment. We chose VirtualHome for the following reasons: 1) The programs provided by VirtualHome are realistic as they are collected through crowdsourcing~\cite{puig2018virtualhome}. Moreover, videos are challenging since they contain various actions and interactions with objects. We define ``action + object'' in the program as an atomic action in our dataset.  
2) Each program contains a sequence of atomic actions and each atomic action can be precisely localized in the rendered videos. Other useful information, such as semantic segmentation maps with object class labels, is provided as well. 3) Recording from static third-person views is possible. We place multiple cameras inside a room to record household activities undertaken by a virtual agent. Our generated dataset consists of a total number of 185 different videos for 18 different activities using the programs provided by \cite{puig2018virtualhome} without huge modifications. These activities happen inside four rooms: the living room, kitchen, bedroom, and bathroom. There are $178$ different atomic actions. Each video, lasting up to $32$ seconds, includes on average $15$ atomic actions and $2.8$ interacting objects.

We used Tobii Spectrum Eye Tracker with a sampling rate of 1200 Hz to collect human eye movement data. Videos were displayed on the screen with a resolution of $1920 \times 1080$ pixels. Participants were seated at a distance of approximately 60 cm in front of the screen. 
For each video, we collected participants' eye movement data while they watched the VirtualHome videos. We recruited 13 participants (8 males and 5 females). Each participant viewed a complete round of the dataset within three sessions (about 15 minutes/session). To ensure that participants paid attention to the critical spatial-temporal information, we asked them to choose the class that best matched the video from the total 18 activity classes. This question made participants stay attentive and was used as an attention check to filter out the gaze data with low accuracy. We kept the data with an accuracy above $85\%$ (one participant with an accuracy below $55\%$ was excluded), resulting in $1310$ videos with human gaze data ($996$ for training and $314$ for testing). We divided the data based on activities and camera angles, meaning that in the test and train datasets, there are no samples with identical activity and camera angle pairs. We preprocessed the raw gaze data and kept only gaze fixations by using the Velocity-Threshold Identification algorithm~\cite{olsen2012tobii}, resulting in one gaze fixation in each frame.
More details of the dataset can be found in the Appendix \ref{sec:app-dataset}. 
\section{Methodology}
\label{sec:method}
In this section, we introduce the problem setup, followed by the details of our proposed method for activity recognition and action anticipation. 

\subsection{Problem Setup}
\label{sec:problem setup}
Given a video sequence $\mathcal{I}=\{I_1, I_2, \dots, I_T\}$ depicting activity $y$ that includes $N$ atomic actions $\{a_1, a_2, \dots,  a_{N}\}$ in total, the model is only allowed to see a part of the video $\mathcal{I'}=\{I_1, I_2, \dots, I_{K}\}$, i.e., the frames until $I_{K}$ covering the atomic actions until $a_{j}$. 
The task is to recognize the overall activity $y$ and predict the remaining actions $\{a_{k+1}, ..., a_N\}$ that are necessary to complete the activity. 
The model only has access to frames in the partial video $\mathcal{I'}$ and it outputs the recognized activity $\hat{y}$ as well as a sequence of predicted atomic actions $\{\hat{a}_{k+1}, \hat{a}_{k+2}, ..., \hat{a}_N\}$ in order to complete the activity $\hat{y}$. Note that $N$ denotes the total number of atomic actions that are included in the activity and it varies in different activities. 

\subsection{\method}
\label{sec:method-gvsf}
As graphical models show their advantages in representing spatial and temporal information in videos or structured relations~\cite{li2021representing, zhou2021graph, jing2020visual,chen2022multi,hussein2019videograph}, we propose to obtain graph representations for videos, where nodes represent temporal-spatial information from keyframes and edges equipped with semantic features indicate relations between them. Based on visual-semantic embeddings from graphs, our framework is trained for activity recognition and atomic action prediction.
\subsubsection{Video to graph.}
Given a partial video clip with $K$ frames $\mathcal{I'}=\{I_1, I_2, ..., I_K\}$ along with the eye gaze location $\mathcal{F'} = \{(x_1,y_1), $ \\ $ (x_2,y_2), ...,(x_K,y_K)\}$, we construct a graph $(\mathcal{V}, \mathcal{E})$. 
We consider the gazed-at content of humans as an important visual representation of the activity and capturing the spatial-temporal changes of a video. Hence, we crop a frame $I_t$ centered at the fixation point $(x_t, y_t)$ with the crop size $B$ to get 
image patch $\mathcal{C}^{(x_t, y_t)} \in \mathbb{R}^{B \times B \times 3}$, and its embedding is used as a node feature in the graph denoted as $v_t \in \mathbb{R}^{d_1}$. However, relying solely on visual embeddings can lead to a homogeneous representation, as frames often share similar backgrounds. We thus deploy an object detector to add object-relevant information. 
The identified objects are encoded as edge attributes $e_{i,j} \in \mathbb{R}^{d_2}$ to capture the context information, and $i,j$ indicate the index of source and target nodes that the edge connects. 

Concretely, a visual encoder $f(\cdot)$ is deployed to encode the cropped images $v_t = f(\mathcal{C}^{(x_t, y_t)})$. To avoid redundancy and obtain better representations of past video sequences, we group similar nodes in the graph. To achieve this, we compare the cosine similarity of the current frame embedding $v_t$ with all existing node embeddings. 
If the highest similarity score is smaller than a threshold $\rho$, a new node will be created using $v_t$ as its feature and the edge between the previous node and the new node will be added. Otherwise, the patch will be merged into an existing node that has the highest similarity, and an edge will be added between these two nodes. The algorithm goes through all frames in a video and assigns each frame to a node in the graph. As for edge attributes $e_{i,j}$ in the graph, we define them as the concatenation of the fixated object information of the nodes they connect. In this manner, the edge can serve as hints depending on the activity being performed. For example, interacting with objects such as a ``cellphone'' or a ``remote control'' may have similar visual embeddings as they only occupy a small portion of the frame and share similar visual appearances. However, these objects have distinct semantic meanings, and their labels improve the classification of the activity objectives. Moreover, the transition from node $i$ to $j$ represented by the pair of objects reveals action information underlying these two nodes. For example, if an edge connects a node with "whipped cream" and "fridge", it is easier for the model to deduce that there is an action "walk to the fridge" between the two nodes. Specifically, an object detector is employed to detect the object inside the image patch. A semantic encoder (such as \texttt{Word2Vec}) provides the word embeddings using the object class name. An edge attribution $e_{i,j}$ is obtained by concatenating these embeddings of its connecting nodes. 
Details in the algorithm as well as the general object classifier can be found in the Appendix \ref{sec:app-implementation}. 

\subsubsection{Hierarchical classification.}
Once the graph $\mathcal{G}$ for a video is constructed, a node $v_i$ in $\mathcal{G}$ acquires activity context by incorporating features from its neighbor nodes and edges. We use the Edge-Conditioned Convolution (ECC) \cite{simonovsky2017dynamic} formalized as follows: 
\begin{align}
        v^l_i &= \frac{1}{|\mathcal{N}(i)|} \sum_{j \in \mathcal{N}(i)} h^l(e_{i,j}; w^l) \cdot v^{l-1}_j +b^l \\
        &= \frac{1}{|\mathcal{N}(i)|} \sum_{j \in \mathcal{N}(i)} \Theta^l_{i,j} \cdot v^{l-1}_j  + b^l,
\end{align}
where $l$ indicates the layer index, and $\mathcal{N}(i)$ is the set of neighbor nodes of node $i$, which is defined as all adjacent nodes including $i$ itself (self-loop)~\cite{simonovsky2017dynamic}. $h^l$ is parameterized by trainable weights $w^l$, which takes edge attribute $e_{i,j}$ as input and outputs parameters $\Theta_{i,j}^l$ used to compute the weights for edges. Learnable bias is depicted as $b^l$. The representation $v_i$ after several ECC layers encodes visual and object information of a node enriched with context from its neighbor nodes, which enables the model to learn atomic action patterns in different activity goals. For example, the nodes encoded with the contextual information of ``grab the glass'' exist in the activity ``put the glass in the cabinet'', while the nodes of ``type the keyboard'' in ``work on the computer''. To represent the complete viewed video clip, averaged node features are used:
\begin{equation}
  v_{G} = \frac{1}{|\mathcal{V}|}\sum_{vi \in \mathcal{V}}v_i,
\end{equation}
where $\mathcal{V}$ represents the set of all nodes in the graph $\mathcal{G}$.
This video representation is used to realize the downstream tasks of activity recognition and atomic prediction. For activity recognition, an MLP (multi-layer perceptron) takes $v_G$ as input and predicts the activity class, formulated as $\hat{y} = \sigma(\text{MLP}(v_G))$, where $\sigma$ denotes the activation function. 
Inspired by how humans attempt to solve a task by first setting overall targets and then plan for subactions conditioned on specific targets~\cite{puig2020watch,zhao2021tnt}, we propose to use the idea of hierarchical classification for making predictions of future atomic actions. Concretely, two LSTM layers are employed since a sequence of atomic actions is necessary to accomplish the activity goal. The predicted action sequence is denoted as $\{\hat{a}_{j}\} = \sigma{\text{LSTM}(v_G \oplus \hat{y}_c)}$, where $j$ is the index of the sequence and $\oplus$ represents the concatenation operation. Our model is jointly trained for two tasks, with the cross-entropy loss being used for each task as follows:
\begin{equation}
\label{eq:loss}
    \mathcal{L} = -\sum_{c=1}^{M_1}(y_c\cdot \log(\hat{y}_c)) - \sum_{j=1}^{N-k} \sum_{c=1}^{M_2}(a_{j,c}\cdot \log(\hat{a}_{j,c})),
\end{equation}
where $M_1$ and $M_2$ are the numbers of classes in activity recognition and atomic action prediction, respectively. $N$ depicts the length of the atomic action sequence and $k$ is the length of the sequence given in the viewed video. We omit the sample index used in batch training for simplicity.

\section{Experiments}
\label{sec:experiments}
\subsection{Evaluation metrics}
\label{sec:eval metrics}
For activity recognition, we use classification accuracy as the evaluation metric. For atomic action prediction, we adopt the Intersection over Union (IoU) and the normalized Levenshtein distance (Leven.) to evaluate the quality of the predicted sequences. The former does not consider the order of the sequence but the latter does. Specifically, IoU is calculated between the ground-truth action sequence $A$ and the predicted sequence $\hat{A}$.

As the final goal is to help the human agent complete the whole activity, we evaluate the quality of predicted action sequences by testing the completion of the task in the VirtualHome platform. Concretely, for each test sample, we generate a complete program by concatenating the predicted sequence $\hat{A}$ to the sequence of the viewed video and test this program on the VirualHome platform to see whether the activity could be successfully executed. 
We report the overall success rate (SR), which is the number of successfully executed programs divided by the total number of test samples. We use $70\%$ of each video as input $\mathcal{I'}$ with the number of remaining action sequences being variable and the end-of-sequence token being predicted in the following experiments. Experiments on other settings, i.e., $[50\%, 90\%]$, can be found in the Appendix \ref{app-more_quantitative}. For the purpose of reproducibility, Appendix \ref{app-hyper-parameter} and \ref{sec:app-implementation} contain implementation details concerning the experimental setup.

\subsection{Experimental Results}
\label{sec:experiments-activity}

\subsubsection{Comparison with state-of-the-art.}
We compare our complete framework with four other state-of-the-art methods: I3D~\cite{carreira2017quo}, Ego-Topo~\cite{nagarajan2020ego}, IGA~\cite{min2021integrating}, VideoGraph~\cite{hussein2019videograph}. 
As all these models are only able to recognize activities or predict atomic actions as a multi-label classification task, i.e., there is no order of the atomic actions, we add the LSTM layer to enable the action prediction. 
We train the modified model with our data and use the loss in Equation \ref{eq:loss}.
I3D is an advanced 3DCNN model for activity recognition tasks, which serves as the backbone in both VideoGraph and IGA, facilitating the extraction of visual embeddings. In VideoGraph, a graph (with a fixed number of neighbor nodes and edges that a node can have) is first defined. Subsequently, a network is trained to convert video clips from a complete video into the aforementioned graph. 
Ego-Topo employs a methodology highly reminiscent of ours for distilling graphs from videos. Specifically, it obtains nodes by assessing frame similarities, which also guides the establishment of edges, capturing the temporal sequence. In comparison to Ego-Topo, our method goes a step further by incorporating fixated object information into the attributes of the edges that connect the nodes. Besides, we also list human performance in the activity classification task in Table \ref{tab:comparison}, where humans are $94\%$ accurate. 
As Ego-Topo and our model both use human gaze, they achieve significantly better performance than other compared methods such as IGA, I3D, or VideoGraph. Ego-Topo achieves only 0.19 in the final success rate of using the predicted atomic actions while ours achieves 0.27 in the success rate.
These results highlight that our proposed method is able to achieve better performance in both activity recognition and action prediction.

\subsubsection{Ablation study.}

In this section, we conduct a comprehensive ablation study to verify the advantages of two components in our {\method} method: \textbf{A1}: incorporation of human gaze-based attention knowledge; \textbf{A2}: benefits of conditioning the action prediction task on the overall activity.

\textbf{A1}: Table \ref{tab:ablation} (upper part) demonstrates the benefits of incorporating human attention in our model compared to the baselines. For the ``random fixation'' baseline, a fixation is randomly sampled in each frame, while ``no fixation''  means that all frames in videos are used as input and no gaze fixation is given. Fixation sequences in the ``random scanpath'' baseline exhibit human gaze behavior but do not match the visual content. Specifically, we collect gaze fixation data for all videos in the VirtualHome Video Dataset from an additional participant and randomly assign a scanpath to each video. From the results, we see that random fixation or scanpath negatively impacts performance because essential information can be lost. In contrast, using the whole video and no cropping on frames preserves all pertinent information, which surpasses the baseline with random fixation.
However, our model integrated with human attention outperforms the baseline by a large margin, increasing by $0.28$ in activity recognition accuracy and by $0.18$ in IoU for atomic action prediction, demonstrating the effectiveness of gaze guidance in both action anticipation and recognition tasks.

\begin{table}[ht]
\caption{Comparison of our framework with other methods. \\ $^*$ indicates that human gaze is used. }
\centering
\begin{tabular}{c|c|c|c|c}
\toprule[1pt]
    & Acc. $\uparrow$ & IoU $\uparrow$ & Leven. $\downarrow$ &  SR $\uparrow$\\\hline
Human & \textbf{0.94} & - & - & - \\\hline
I3D & 0.12   &  0.03  & 0.79    & 0.06 \\ \hline
VideoGraph & 0.09  &    0.09  & 0.71   & 0.08 \\ \hline
IGA &  0.27  & 0.21 &   0.75 &  0.14     \\ \hline
Ego-Topo$^*$ &  0.54 &  0.26 & 0.63 &  0.19 \\
\midrule
\midrule
Ours$^*$  & 0.61 & \textbf{0.35} & \textbf{0.51} & \textbf{0.27}  \\ 
\bottomrule[1pt]
\end{tabular}
\label{tab:comparison}
\end{table}

\begin{table}[ht]
\caption{\textbf{Upper}: Influence of human gaze fixation in activity recognition and the atomic action prediction tasks (\textbf{A1}). \textbf{Bottom}: Influence of incorporating activity recognition in atomic action prediction (\textbf{A2}).}
\centering
\begin{tabular}{c|c|c|c}
\toprule[1pt]
    &  Acc. $\uparrow$ & IoU $\uparrow$ & Leven. $\downarrow$ \\\hline
No Fixation (full video) & 0.33 & 0.17 & 0.74 \\\hline
Random Fixation &  0.30  & 0.14     &  0.87     \\ \hline
Random Scanpath & 0.26   &  0.21     &  0.73     \\ 
\midrule[.8pt]
\midrule[.8pt]
W/o Activity Recog.  &  -  &  0.30    &    0.61  \\ \hline
With Activity Recog. &  0.61    &   0.33  &   0.55    \\
\midrule[.8pt]
\midrule[.8pt]
Ours &\textbf{0.61} & \textbf{0.35} &  \textbf{0.51} \\
\bottomrule[1pt]
\end{tabular}
\label{tab:ablation}
\end{table}

\textbf{A2}: Understanding the overall goal benefits the model in designing the remaining steps to achieve this goal. We validate this effect using results in Table \ref{tab:ablation} (bottom part). The baseline ``w/o activity recog.'' is the model trained only with the loss of action prediction, while ``with activity recog.'' is the one trained with the loss in Equation~\ref{eq:loss} but without using the conditioned action prediction (i.e., hierarchical pred.). We observe that co-training with two parts of losses increases the performance in action prediction compared to the results without the supervision of overall activity goals. As shown in the table, it improves the Levenshtein distance from $0.61$ to $0.55$. Using the proposed hierarchical classification achieves the best performance, especially in improving the Levenshtein distance. This reveals that conditioning on the overall goal significantly helps the model discover the intermediate steps required to accomplish it.

\subsubsection{Qualitative Results.}
\begin{figure}[ht]
    \centering
    \includegraphics[width=0.780\linewidth]{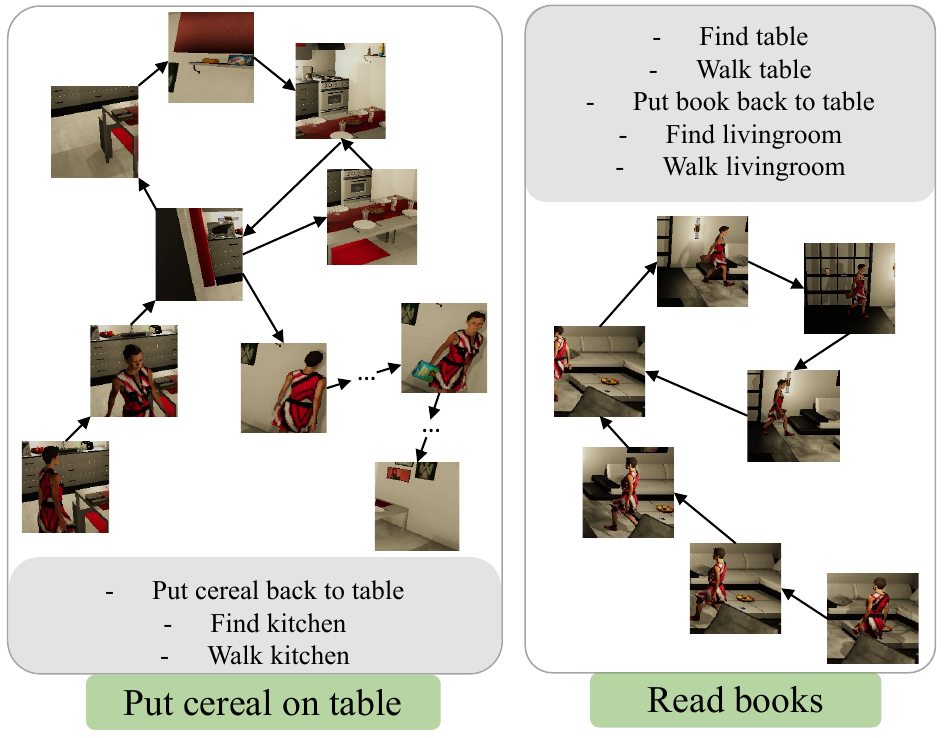}
    \caption{Two examples of predicted actions using our model. The predicted action sequence is shown below the graph. Graphs are visualized based on the viewed video, where some nodes are omitted with ``$\dots$'' for a clearer view. The intention is given at the bottom. }
    \Description{Two examples of predicted actions using our model. The predicted action sequence is shown below the graph. Graphs are visualized based on the viewed video, where some nodes are omitted with ``$\dots$'' for a clearer view. The intention is given at the bottom. }
    \label{fig:quali}
\end{figure}

We present two examples of action prediction and recognition using our model in Figure~\ref{fig:quali}. The graphs are constructed using our human gaze fixations on the input $70\%$ of the video. The left example is ``put cereal on the table''. In this graph, the first node is at the left bottom and the last node is at the right bottom, the first two fixations fall on the agent, followed by several nodes exploring the environment. For instance, the fixation starts to look at the microwave, the table, and the cereal on the shelf. Our object detector successfully recognizes these objects in those nodes and the graph is integrated with this information. 
Then, the agent grabs the cereal and turns towards the table. Interestingly, the last fixation falls on the corner of the table, indicating the correct direction for the agent's movement. The predicted action sequence fulfills the activity goal, where the next atomic is ``put cereal back on table''.

The second example happens in the living room, where the agent intends to read a book and put the book back on the shelf after reading for a while. From the graph, we see that the first $70\%$ of the video contains the actions of the agent walking to the bookshelf, grabbing the book, and then walking to the sofa. Our object detector labels the nodes most of the time with ``sofa'', which is relevant to the environment where this activity happens. It is worth mentioning that our graph is able to establish the spatial relation between nodes. For instance, the fourth node in the time sequence is the area of the sofa. After finding the book, the node reconnects to this node, as it appears the human agent will sit on the sofa to read the book. The future atomic action after reading the book is that the agent will go to the table and put the book back on the table. More qualitative examples can be found in the Appendix~\ref{app-more-qualitative}.

\section{Conclusion}
In this paper, we tackle a challenging human action anticipation problem, where a model views humans performing a partial long-term activity, then recognizes the human's intention (activity, goal) and predicts the necessary atomic actions to achieve this final goal. We propose a framework, \method, to address this challenge by transforming video into graphs encoded with essential spatial-temporal information and object hints based on human gaze guidance. A graph neural network is utilized to predict the atomic action sequence conditioned on the recognized human intention.
Our experimental results demonstrate the advantages of our framework compared to other methods, as our model can better simulate how humans perceive and process the content of the video, highlighting its potential in human intention understanding. In our future work, we plan to evaluate the applicability of our proposed model on real-world datasets, gauging its ability to perform effectively across a broader spectrum of scenarios.

\begin{acks}
We are deeply grateful to IT-Stiftung Esslingen for their generous support of our hardware lab. 
\end{acks}

\bibliographystyle{ACM-Reference-Format}
\bibliography{ETRA_action}


\begin{thebibliography}{33}


\ifx \showCODEN    \undefined \def \showCODEN     #1{\unskip}     \fi
\ifx \showDOI      \undefined \def \showDOI       #1{#1}\fi
\ifx \showISBNx    \undefined \def \showISBNx     #1{\unskip}     \fi
\ifx \showISBNxiii \undefined \def \showISBNxiii  #1{\unskip}     \fi
\ifx \showISSN     \undefined \def \showISSN      #1{\unskip}     \fi
\ifx \showLCCN     \undefined \def \showLCCN      #1{\unskip}     \fi
\ifx \shownote     \undefined \def \shownote      #1{#1}          \fi
\ifx \showarticletitle \undefined \def \showarticletitle #1{#1}   \fi
\ifx \showURL      \undefined \def \showURL       {\relax}        \fi
\providecommand\bibfield[2]{#2}
\providecommand\bibinfo[2]{#2}
\providecommand\natexlab[1]{#1}
\providecommand\showeprint[2][]{arXiv:#2}

\bibitem[Aronson et~al\mbox{.}(2018)]%
        {aronson2018eye}
\bibfield{author}{\bibinfo{person}{Reuben~M Aronson}, \bibinfo{person}{Thiago
  Santini}, \bibinfo{person}{Thomas~C K{\"u}bler}, \bibinfo{person}{Enkelejda
  Kasneci}, \bibinfo{person}{Siddhartha Srinivasa}, {and}
  \bibinfo{person}{Henny Admoni}.} \bibinfo{year}{2018}\natexlab{}.
\newblock \showarticletitle{Eye-hand behavior in human-robot shared
  manipulation}. In \bibinfo{booktitle}{\emph{Proceedings of the 2018 ACM/IEEE
  International Conference on Human-Robot Interaction}}.
  \bibinfo{pages}{4--13}.
\newblock


\bibitem[Braunagel et~al\mbox{.}(2017)]%
        {braunagel2017ready}
\bibfield{author}{\bibinfo{person}{Christian Braunagel},
  \bibinfo{person}{Wolfgang Rosenstiel}, {and} \bibinfo{person}{Enkelejda
  Kasneci}.} \bibinfo{year}{2017}\natexlab{}.
\newblock \showarticletitle{Ready for take-over? A new driver assistance system
  for an automated classification of driver take-over readiness}.
\newblock \bibinfo{journal}{\emph{IEEE Intelligent Transportation Systems
  Magazine}} \bibinfo{volume}{9}, \bibinfo{number}{4} (\bibinfo{year}{2017}),
  \bibinfo{pages}{10--22}.
\newblock


\bibitem[Carreira and Zisserman(2017)]%
        {carreira2017quo}
\bibfield{author}{\bibinfo{person}{Joao Carreira} {and} \bibinfo{person}{Andrew
  Zisserman}.} \bibinfo{year}{2017}\natexlab{}.
\newblock \showarticletitle{Quo vadis, action recognition? a new model and the
  kinetics dataset}. In \bibinfo{booktitle}{\emph{proceedings of the IEEE
  Conference on Computer Vision and Pattern Recognition}}.
  \bibinfo{pages}{6299--6308}.
\newblock


\bibitem[Chen and Li(2022)]%
        {chen2022multi}
\bibfield{author}{\bibinfo{person}{Sijia Chen} {and} \bibinfo{person}{Baochun
  Li}.} \bibinfo{year}{2022}\natexlab{}.
\newblock \showarticletitle{Multi-Modal Dynamic Graph Transformer for Visual
  Grounding}. In \bibinfo{booktitle}{\emph{Proceedings of the IEEE/CVF
  Conference on Computer Vision and Pattern Recognition}}.
  \bibinfo{pages}{15534--15543}.
\newblock


\bibitem[Damen et~al\mbox{.}(2018)]%
        {damen2018scaling}
\bibfield{author}{\bibinfo{person}{Dima Damen}, \bibinfo{person}{Hazel
  Doughty}, \bibinfo{person}{Giovanni~Maria Farinella}, \bibinfo{person}{Sanja
  Fidler}, \bibinfo{person}{Antonino Furnari}, \bibinfo{person}{Evangelos
  Kazakos}, \bibinfo{person}{Davide Moltisanti}, \bibinfo{person}{Jonathan
  Munro}, \bibinfo{person}{Toby Perrett}, \bibinfo{person}{Will Price},
  {et~al\mbox{.}}} \bibinfo{year}{2018}\natexlab{}.
\newblock \showarticletitle{Scaling egocentric vision: The epic-kitchens
  dataset}. In \bibinfo{booktitle}{\emph{Proceedings of the European Conference
  on Computer Vision (ECCV)}}. \bibinfo{pages}{720--736}.
\newblock


\bibitem[Hussein et~al\mbox{.}(2019)]%
        {hussein2019videograph}
\bibfield{author}{\bibinfo{person}{Noureldien Hussein},
  \bibinfo{person}{Efstratios Gavves}, {and} \bibinfo{person}{Arnold~WM
  Smeulders}.} \bibinfo{year}{2019}\natexlab{}.
\newblock \showarticletitle{Videograph: Recognizing minutes-long human
  activities in videos}.
\newblock \bibinfo{journal}{\emph{arXiv preprint arXiv:1905.05143}}
  (\bibinfo{year}{2019}).
\newblock


\bibitem[Jing et~al\mbox{.}(2020)]%
        {jing2020visual}
\bibfield{author}{\bibinfo{person}{Chenchen Jing}, \bibinfo{person}{Yuwei Wu},
  \bibinfo{person}{Mingtao Pei}, \bibinfo{person}{Yao Hu},
  \bibinfo{person}{Yunde Jia}, {and} \bibinfo{person}{Qi Wu}.}
  \bibinfo{year}{2020}\natexlab{}.
\newblock \showarticletitle{Visual-semantic graph matching for visual
  grounding}. In \bibinfo{booktitle}{\emph{Proceedings of the 28th ACM
  International Conference on Multimedia}}. \bibinfo{pages}{4041--4050}.
\newblock


\bibitem[Karargyris et~al\mbox{.}(2021)]%
        {karargyris2021creation}
\bibfield{author}{\bibinfo{person}{Alexandros Karargyris},
  \bibinfo{person}{Satyananda Kashyap}, \bibinfo{person}{Ismini Lourentzou},
  \bibinfo{person}{Joy~T Wu}, \bibinfo{person}{Arjun Sharma},
  \bibinfo{person}{Matthew Tong}, \bibinfo{person}{Shafiq Abedin},
  \bibinfo{person}{David Beymer}, \bibinfo{person}{Vandana Mukherjee},
  \bibinfo{person}{Elizabeth~A Krupinski}, {et~al\mbox{.}}}
  \bibinfo{year}{2021}\natexlab{}.
\newblock \showarticletitle{Creation and validation of a chest X-ray dataset
  with eye-tracking and report dictation for AI development}.
\newblock \bibinfo{journal}{\emph{Scientific data}} \bibinfo{volume}{8},
  \bibinfo{number}{1} (\bibinfo{year}{2021}), \bibinfo{pages}{92}.
\newblock


\bibitem[Kraus et~al\mbox{.}(2022)]%
        {kraus2022kurt}
\bibfield{author}{\bibinfo{person}{Matthias Kraus}, \bibinfo{person}{Nicolas
  Wagner}, \bibinfo{person}{Wolfgang Minker}, \bibinfo{person}{Ankita Agrawal},
  \bibinfo{person}{Artur Schmidt}, \bibinfo{person}{Pranav~Krishna Prasad},
  {and} \bibinfo{person}{Wolfgang Ertel}.} \bibinfo{year}{2022}\natexlab{}.
\newblock \showarticletitle{KURT: A Household Assistance Robot Capable of
  Proactive Dialogue}. In \bibinfo{booktitle}{\emph{2022 17th ACM/IEEE
  International Conference on Human-Robot Interaction (HRI)}}. IEEE,
  \bibinfo{pages}{855--859}.
\newblock


\bibitem[Li et~al\mbox{.}(2021)]%
        {li2021representing}
\bibfield{author}{\bibinfo{person}{Dong Li}, \bibinfo{person}{Zhaofan Qiu},
  \bibinfo{person}{Yingwei Pan}, \bibinfo{person}{Ting Yao},
  \bibinfo{person}{Houqiang Li}, {and} \bibinfo{person}{Tao Mei}.}
  \bibinfo{year}{2021}\natexlab{}.
\newblock \showarticletitle{Representing videos as discriminative sub-graphs
  for action recognition}. In \bibinfo{booktitle}{\emph{Proceedings of the
  IEEE/CVF Conference on Computer Vision and Pattern Recognition}}.
  \bibinfo{pages}{3310--3319}.
\newblock


\bibitem[Li et~al\mbox{.}(2018)]%
        {li2018eye}
\bibfield{author}{\bibinfo{person}{Yin Li}, \bibinfo{person}{Miao Liu}, {and}
  \bibinfo{person}{James~M Rehg}.} \bibinfo{year}{2018}\natexlab{}.
\newblock \showarticletitle{In the eye of beholder: Joint learning of gaze and
  actions in first person video}. In \bibinfo{booktitle}{\emph{Proceedings of
  the European conference on computer vision (ECCV)}}.
  \bibinfo{pages}{619--635}.
\newblock


\bibitem[Li et~al\mbox{.}(2015)]%
        {li2015delving}
\bibfield{author}{\bibinfo{person}{Yin Li}, \bibinfo{person}{Zhefan Ye}, {and}
  \bibinfo{person}{James~M Rehg}.} \bibinfo{year}{2015}\natexlab{}.
\newblock \showarticletitle{Delving into egocentric actions}. In
  \bibinfo{booktitle}{\emph{Proceedings of the IEEE conference on computer
  vision and pattern recognition}}. \bibinfo{pages}{287--295}.
\newblock


\bibitem[Mikolov et~al\mbox{.}(2013)]%
        {mikolov2013efficient}
\bibfield{author}{\bibinfo{person}{Tomas Mikolov}, \bibinfo{person}{Kai Chen},
  \bibinfo{person}{Greg Corrado}, {and} \bibinfo{person}{Jeffrey Dean}.}
  \bibinfo{year}{2013}\natexlab{}.
\newblock \showarticletitle{Efficient estimation of word representations in
  vector space}.
\newblock \bibinfo{journal}{\emph{arXiv preprint arXiv:1301.3781}}
  (\bibinfo{year}{2013}).
\newblock


\bibitem[Min and Corso(2021)]%
        {min2021integrating}
\bibfield{author}{\bibinfo{person}{Kyle Min} {and} \bibinfo{person}{Jason~J
  Corso}.} \bibinfo{year}{2021}\natexlab{}.
\newblock \showarticletitle{Integrating human gaze into attention for
  egocentric activity recognition}. In \bibinfo{booktitle}{\emph{Proceedings of
  the IEEE/CVF Winter Conference on Applications of Computer Vision}}.
  \bibinfo{pages}{1069--1078}.
\newblock


\bibitem[Nagarajan et~al\mbox{.}(2020)]%
        {nagarajan2020ego}
\bibfield{author}{\bibinfo{person}{Tushar Nagarajan}, \bibinfo{person}{Yanghao
  Li}, \bibinfo{person}{Christoph Feichtenhofer}, {and}
  \bibinfo{person}{Kristen Grauman}.} \bibinfo{year}{2020}\natexlab{}.
\newblock \showarticletitle{Ego-topo: Environment affordances from egocentric
  video}. In \bibinfo{booktitle}{\emph{Proceedings of the IEEE/CVF Conference
  on Computer Vision and Pattern Recognition}}. \bibinfo{pages}{163--172}.
\newblock


\bibitem[Olsen(2012)]%
        {olsen2012tobii}
\bibfield{author}{\bibinfo{person}{Anneli Olsen}.}
  \bibinfo{year}{2012}\natexlab{}.
\newblock \showarticletitle{The Tobii I-VT fixation filter}.
\newblock \bibinfo{journal}{\emph{Tobii Technology}}  \bibinfo{volume}{21}
  (\bibinfo{year}{2012}), \bibinfo{pages}{4--19}.
\newblock


\bibitem[Panetta et~al\mbox{.}(2019)]%
        {panetta2019software}
\bibfield{author}{\bibinfo{person}{Karen Panetta}, \bibinfo{person}{Qianwen
  Wan}, \bibinfo{person}{Aleksandra Kaszowska}, \bibinfo{person}{Holly~A
  Taylor}, {and} \bibinfo{person}{Sos Agaian}.}
  \bibinfo{year}{2019}\natexlab{}.
\newblock \showarticletitle{Software architecture for automating cognitive
  science eye-tracking data analysis and object annotation}.
\newblock \bibinfo{journal}{\emph{IEEE Transactions on Human-Machine Systems}}
  \bibinfo{volume}{49}, \bibinfo{number}{3} (\bibinfo{year}{2019}),
  \bibinfo{pages}{268--277}.
\newblock


\bibitem[Pham et~al\mbox{.}(2017)]%
        {pham2017evaluating}
\bibfield{author}{\bibinfo{person}{Thi Xuan~Ngan Pham}, \bibinfo{person}{Kotaro
  Hayashi}, \bibinfo{person}{Christian Becker-Asano},
  \bibinfo{person}{Sebastian Lacher}, {and} \bibinfo{person}{Ikuo Mizuuchi}.}
  \bibinfo{year}{2017}\natexlab{}.
\newblock \showarticletitle{Evaluating the usability and users' acceptance of a
  kitchen assistant robot in household environment}. In
  \bibinfo{booktitle}{\emph{2017 26th IEEE international symposium on robot and
  human interactive communication (RO-MAN)}}. IEEE, \bibinfo{pages}{987--992}.
\newblock


\bibitem[Puig et~al\mbox{.}(2018)]%
        {puig2018virtualhome}
\bibfield{author}{\bibinfo{person}{Xavier Puig}, \bibinfo{person}{Kevin Ra},
  \bibinfo{person}{Marko Boben}, \bibinfo{person}{Jiaman Li},
  \bibinfo{person}{Tingwu Wang}, \bibinfo{person}{Sanja Fidler}, {and}
  \bibinfo{person}{Antonio Torralba}.} \bibinfo{year}{2018}\natexlab{}.
\newblock \showarticletitle{Virtualhome: Simulating household activities via
  programs}. In \bibinfo{booktitle}{\emph{Proceedings of the IEEE Conference on
  Computer Vision and Pattern Recognition}}. \bibinfo{pages}{8494--8502}.
\newblock


\bibitem[Puig et~al\mbox{.}(2020)]%
        {puig2020watch}
\bibfield{author}{\bibinfo{person}{Xavier Puig}, \bibinfo{person}{Tianmin Shu},
  \bibinfo{person}{Shuang Li}, \bibinfo{person}{Zilin Wang},
  \bibinfo{person}{Yuan-Hong Liao}, \bibinfo{person}{Joshua~B Tenenbaum},
  \bibinfo{person}{Sanja Fidler}, {and} \bibinfo{person}{Antonio Torralba}.}
  \bibinfo{year}{2020}\natexlab{}.
\newblock \showarticletitle{Watch-and-help: A challenge for social perception
  and human-ai collaboration}.
\newblock \bibinfo{journal}{\emph{arXiv preprint arXiv:2010.09890}}
  (\bibinfo{year}{2020}).
\newblock


\bibitem[Radford et~al\mbox{.}(2021)]%
        {radford2021learning}
\bibfield{author}{\bibinfo{person}{Alec Radford}, \bibinfo{person}{Jong~Wook
  Kim}, \bibinfo{person}{Chris Hallacy}, \bibinfo{person}{Aditya Ramesh},
  \bibinfo{person}{Gabriel Goh}, \bibinfo{person}{Sandhini Agarwal},
  \bibinfo{person}{Girish Sastry}, \bibinfo{person}{Amanda Askell},
  \bibinfo{person}{Pamela Mishkin}, \bibinfo{person}{Jack Clark},
  {et~al\mbox{.}}} \bibinfo{year}{2021}\natexlab{}.
\newblock \showarticletitle{Learning transferable visual models from natural
  language supervision}. In \bibinfo{booktitle}{\emph{International conference
  on machine learning}}. PMLR, \bibinfo{pages}{8748--8763}.
\newblock


\bibitem[Rong et~al\mbox{.}(2021)]%
        {rong2021human}
\bibfield{author}{\bibinfo{person}{Yao Rong}, \bibinfo{person}{Wenjia Xu},
  \bibinfo{person}{Zeynep Akata}, {and} \bibinfo{person}{Enkelejda Kasneci}.}
  \bibinfo{year}{2021}\natexlab{}.
\newblock \showarticletitle{Human Attention in Fine-grained Classification}.
\newblock \bibinfo{journal}{\emph{BMVC}} (\bibinfo{year}{2021}).
\newblock


\bibitem[Santella et~al\mbox{.}(2006)]%
        {santella2006gaze}
\bibfield{author}{\bibinfo{person}{Anthony Santella}, \bibinfo{person}{Maneesh
  Agrawala}, \bibinfo{person}{Doug DeCarlo}, \bibinfo{person}{David Salesin},
  {and} \bibinfo{person}{Michael Cohen}.} \bibinfo{year}{2006}\natexlab{}.
\newblock \showarticletitle{Gaze-based interaction for semi-automatic photo
  cropping}. In \bibinfo{booktitle}{\emph{Proceedings of the SIGCHI conference
  on Human Factors in computing systems}}. \bibinfo{pages}{771--780}.
\newblock


\bibitem[Shafti et~al\mbox{.}(2019)]%
        {shafti2019gaze}
\bibfield{author}{\bibinfo{person}{Ali Shafti}, \bibinfo{person}{Pavel Orlov},
  {and} \bibinfo{person}{A~Aldo Faisal}.} \bibinfo{year}{2019}\natexlab{}.
\newblock \showarticletitle{Gaze-based, context-aware robotic system for
  assisted reaching and grasping}. In \bibinfo{booktitle}{\emph{2019
  International Conference on Robotics and Automation (ICRA)}}. IEEE,
  \bibinfo{pages}{863--869}.
\newblock


\bibitem[Simonovsky and Komodakis(2017)]%
        {simonovsky2017dynamic}
\bibfield{author}{\bibinfo{person}{Martin Simonovsky} {and}
  \bibinfo{person}{Nikos Komodakis}.} \bibinfo{year}{2017}\natexlab{}.
\newblock \showarticletitle{Dynamic edge-conditioned filters in convolutional
  neural networks on graphs}. In \bibinfo{booktitle}{\emph{Proceedings of the
  IEEE conference on computer vision and pattern recognition}}.
  \bibinfo{pages}{3693--3702}.
\newblock


\bibitem[Vasconez et~al\mbox{.}(2019)]%
        {vasconez2019human}
\bibfield{author}{\bibinfo{person}{Juan~P Vasconez}, \bibinfo{person}{George~A
  Kantor}, {and} \bibinfo{person}{Fernando A~Auat Cheein}.}
  \bibinfo{year}{2019}\natexlab{}.
\newblock \showarticletitle{Human--robot interaction in agriculture: A survey
  and current challenges}.
\newblock \bibinfo{journal}{\emph{Biosystems engineering}}
  \bibinfo{volume}{179} (\bibinfo{year}{2019}), \bibinfo{pages}{35--48}.
\newblock


\bibitem[Villani et~al\mbox{.}(2018)]%
        {villani2018survey}
\bibfield{author}{\bibinfo{person}{Valeria Villani}, \bibinfo{person}{Fabio
  Pini}, \bibinfo{person}{Francesco Leali}, {and} \bibinfo{person}{Cristian
  Secchi}.} \bibinfo{year}{2018}\natexlab{}.
\newblock \showarticletitle{Survey on human--robot collaboration in industrial
  settings: Safety, intuitive interfaces and applications}.
\newblock \bibinfo{journal}{\emph{Mechatronics}}  \bibinfo{volume}{55}
  (\bibinfo{year}{2018}), \bibinfo{pages}{248--266}.
\newblock


\bibitem[Weber et~al\mbox{.}(2020)]%
        {weber2020distilling}
\bibfield{author}{\bibinfo{person}{Daniel Weber}, \bibinfo{person}{Thiago
  Santini}, \bibinfo{person}{Andreas Zell}, {and} \bibinfo{person}{Enkelejda
  Kasneci}.} \bibinfo{year}{2020}\natexlab{}.
\newblock \showarticletitle{Distilling location proposals of unknown objects
  through gaze information for human-robot interaction}. In
  \bibinfo{booktitle}{\emph{2020 IEEE/RSJ International Conference on
  Intelligent Robots and Systems (IROS)}}. IEEE, \bibinfo{pages}{11086--11093}.
\newblock


\bibitem[Xia et~al\mbox{.}(2020)]%
        {xia2020periphery}
\bibfield{author}{\bibinfo{person}{Ye Xia}, \bibinfo{person}{Jinkyu Kim},
  \bibinfo{person}{John Canny}, \bibinfo{person}{Karl Zipser},
  \bibinfo{person}{Teresa Canas-Bajo}, {and} \bibinfo{person}{David Whitney}.}
  \bibinfo{year}{2020}\natexlab{}.
\newblock \showarticletitle{Periphery-fovea multi-resolution driving model
  guided by human attention}. In \bibinfo{booktitle}{\emph{Proceedings of the
  IEEE/CVF Winter Conference on Applications of Computer Vision}}.
  \bibinfo{pages}{1767--1775}.
\newblock


\bibitem[Xia et~al\mbox{.}(2019)]%
        {xia2019predicting}
\bibfield{author}{\bibinfo{person}{Ye Xia}, \bibinfo{person}{Danqing Zhang},
  \bibinfo{person}{Jinkyu Kim}, \bibinfo{person}{Ken Nakayama},
  \bibinfo{person}{Karl Zipser}, {and} \bibinfo{person}{David Whitney}.}
  \bibinfo{year}{2019}\natexlab{}.
\newblock \showarticletitle{Predicting driver attention in critical
  situations}. In \bibinfo{booktitle}{\emph{ACCV}}. Springer,
  \bibinfo{pages}{658--674}.
\newblock


\bibitem[Zhao et~al\mbox{.}(2021)]%
        {zhao2021tnt}
\bibfield{author}{\bibinfo{person}{Hang Zhao}, \bibinfo{person}{Jiyang Gao},
  \bibinfo{person}{Tian Lan}, \bibinfo{person}{Chen Sun}, \bibinfo{person}{Ben
  Sapp}, \bibinfo{person}{Balakrishnan Varadarajan}, \bibinfo{person}{Yue
  Shen}, \bibinfo{person}{Yi Shen}, \bibinfo{person}{Yuning Chai},
  \bibinfo{person}{Cordelia Schmid}, {et~al\mbox{.}}}
  \bibinfo{year}{2021}\natexlab{}.
\newblock \showarticletitle{Tnt: Target-driven trajectory prediction}. In
  \bibinfo{booktitle}{\emph{Conference on Robot Learning}}. PMLR,
  \bibinfo{pages}{895--904}.
\newblock


\bibitem[Zheng et~al\mbox{.}(2022)]%
        {zheng2022gimo}
\bibfield{author}{\bibinfo{person}{Yang Zheng}, \bibinfo{person}{Yanchao Yang},
  \bibinfo{person}{Kaichun Mo}, \bibinfo{person}{Jiaman Li},
  \bibinfo{person}{Tao Yu}, \bibinfo{person}{Yebin Liu},
  \bibinfo{person}{C~Karen Liu}, {and} \bibinfo{person}{Leonidas~J Guibas}.}
  \bibinfo{year}{2022}\natexlab{}.
\newblock \showarticletitle{Gimo: Gaze-informed human motion prediction in
  context}. In \bibinfo{booktitle}{\emph{European Conference on Computer
  Vision}}. Springer, \bibinfo{pages}{676--694}.
\newblock


\bibitem[Zhou et~al\mbox{.}(2021)]%
        {zhou2021graph}
\bibfield{author}{\bibinfo{person}{Jiaming Zhou}, \bibinfo{person}{Kun-Yu Lin},
  \bibinfo{person}{Haoxin Li}, {and} \bibinfo{person}{Wei-Shi Zheng}.}
  \bibinfo{year}{2021}\natexlab{}.
\newblock \showarticletitle{Graph-based high-order relation modeling for
  long-term action recognition}. In \bibinfo{booktitle}{\emph{Proceedings of
  the IEEE/CVF Conference on Computer Vision and Pattern Recognition}}.
  \bibinfo{pages}{8984--8993}.
\newblock


\end{thebibliography}

\newpage
\clearpage

\appendix
\section{Virtualhome Video Dataset Statistics}
\label{sec:app-dataset}

\begin{figure*}[ht]
     \centering
     \begin{subfigure}[b]{0.44\linewidth}
         \centering
        \includegraphics[width=\linewidth]{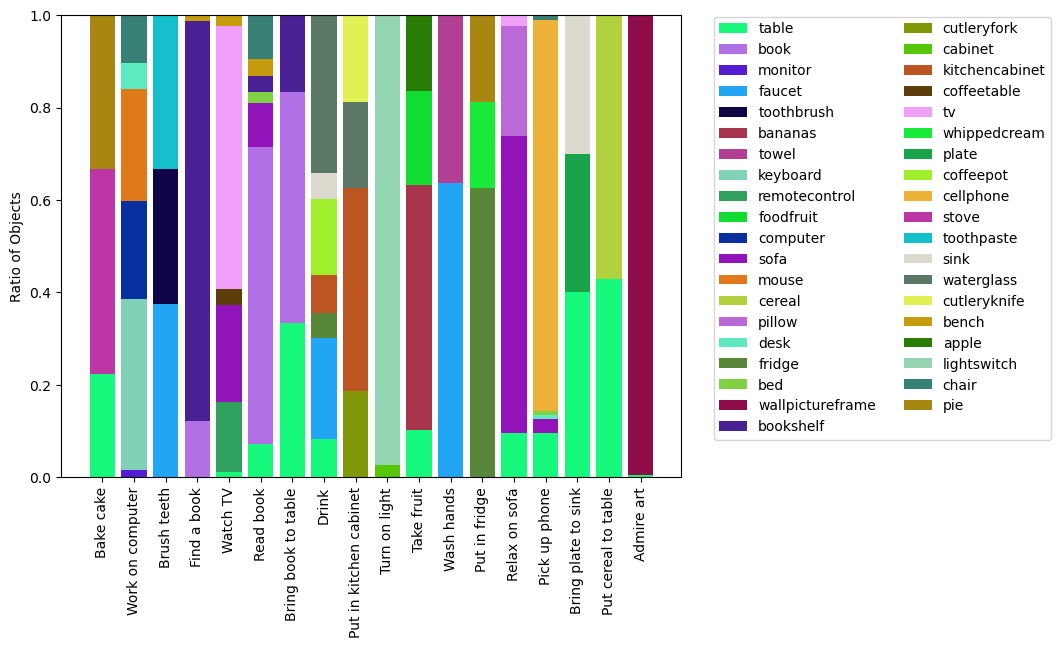}
        \caption{}
        \label{app-fig:objects}
     \end{subfigure}
     \begin{subfigure}[b]{0.44\linewidth}
         \centering
        \includegraphics[width=\linewidth]{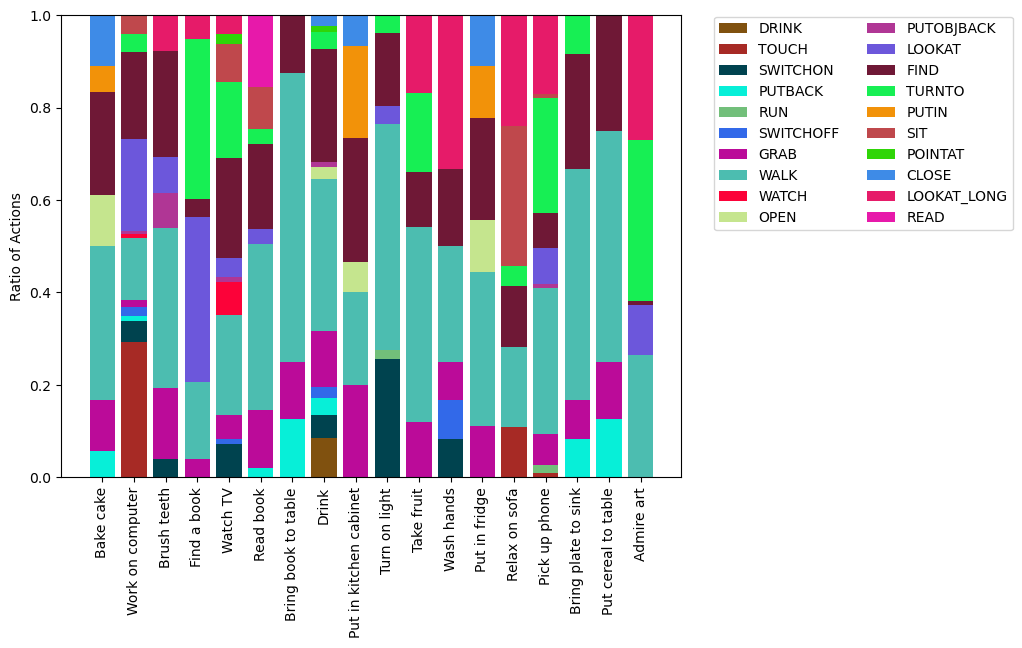}
        \caption{}
        \label{app-fig:action}
     \end{subfigure}
     \\
     \begin{subfigure}[b]{0.5\linewidth}
        \includegraphics[width=\linewidth]{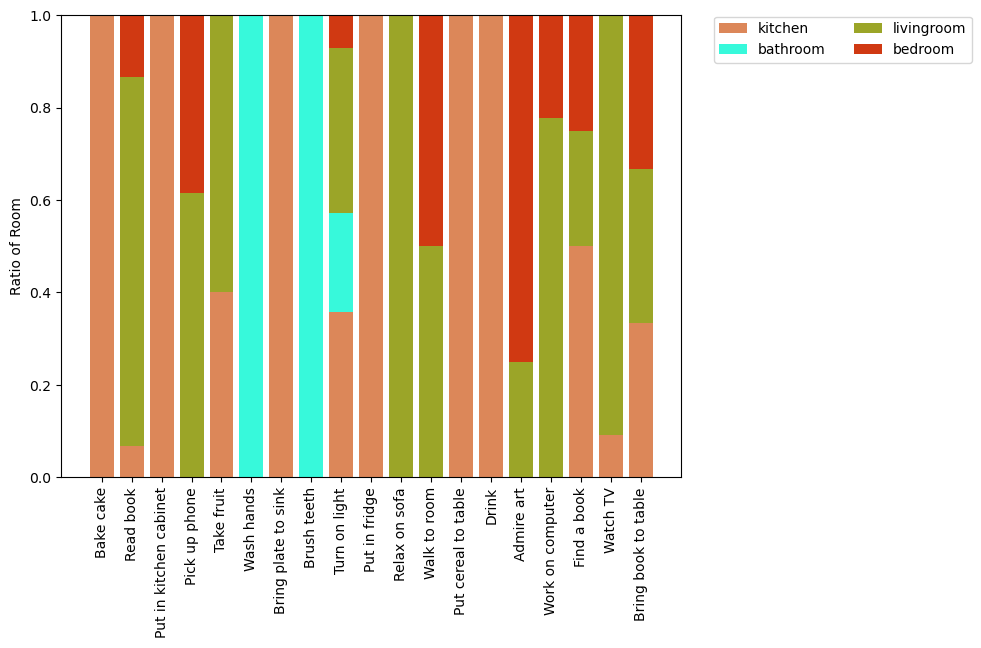}
        \caption{}
        \label{app-fig:room}
     \end{subfigure}
        \caption{Dataset statistics: (a) Distribution of interacting objects for each activity. (b) Distribution of atomic actions for each activity. (c) Distribution rooms for each activity. }
        \Description{Dataset statistics: (a) Distribution of interacting objects for each activity. (b) Distribution of atomic actions for each activity. (c) Distribution rooms for each activity.}
        \vspace{-1em}
\end{figure*}

In this section, we demonstrate the statics of different objects that the agent interacts with, atomic actions, and rooms inside each activity. In Figure~\ref{app-fig:objects}, we see that, for instance, there are seven different objects that appear in the activity "Drink", and most of the time, the agent interacts with the "water glass" in this activity. Figure~\ref{app-fig:action} illustrates the distribution of different atomic actions of an activity. For example, "work on a computer" contains twelve different atomic actions, where "touch" is the most frequent action, because the agent is touching the mouse or keyboard. These activities happen inside four different rooms, and most of the activities contain two rooms, as illustrated in Figure\ref{app-fig:room}.

\section{Hyper-parameter Tuning} \label{app-hyper-parameter}
The whole algorithm of \method{} can be found in  Algorithm \ref{alg:video2graph}.
In this section, we discuss hyper-parameters in our method: (1)  the crop size $B$, which represents the fovea area radius of humans; (2) the threshold $\rho$ determining the similarity for merging visual embeddings into one node.
\begin{algorithm}[ht]
\caption{Video to Graph} 
\label{alg:video2graph}
\begin{algorithmic}[1]
\REQUIRE{A video $\mathcal{I'}=\{I_1, I_2, \dots, I_K\}$; Eye gaze fixation coordinate $\mathcal{F'} = \{(x_1,y_1), (x_2,y_2),\dots,(x_K,y_K)\}$; Crop size $B$; Node similarity threshold $\rho$.}
\ENSURE{A graph $\mathcal{G} = (\mathcal{V}, \mathcal{E})$.}
\STATE{$\mathcal{V}$ $\gets [f(\mathcal{C}^{(x_1, y_1)}]$}: Initialize the node list $\mathcal{V}$ with the first frame.
\STATE {$\mathcal{E}$ $\gets []$}: Initialize the edge $\mathcal{E}$  list with an empty list.
\STATE {previous node index $i \gets 1$.}
\FOR{$t \gets $2 to $K$}                    
    \STATE{$v_t \gets  f(\mathcal{C}^{(x_t, y_t)})$}
    \IF{$\max cos(v_j, v_t) \geq  \rho$, $v_j \in \mathcal{V}$}
    \STATE{$i^* = \argmax_{j} \cos(v_j, v_t)$.} 
    \IF{$i^* \neq i$} 
    \STATE{Add the edge $e_{i,i^*}$ to $\mathcal{E}$.}
    \STATE{\color{lightgray} \# $(i, i^*)$ is the source and target node index.}
    \STATE{$i \gets i^*$.}
    \ENDIF
    \STATE{Add a new node $v_t$ to $\mathcal{V}$.}
    \STATE{\color{lightgray} \# Detect the fixated object of this node.}    
    \STATE{$i^* \gets$ $|\mathcal{V}|$.}
    \STATE{Add the edge $e_{i,i^*}$ to $\mathcal{E}$.}
    \STATE{\color{lightgray} \# $(i, i^*)$ is the source and target node index.}
     \STATE{$i \gets i^*$.}
 \ENDIF
\ENDFOR
\STATE \RETURN {$\mathcal{G} = (\mathcal{V}, \mathcal{E})$}
\end{algorithmic}

\end{algorithm}

The input frame consists of the complete view of a room, but the agent operates only in a local area within the room. Thus, cropping the image based on the fovea area is beneficial for the model as it reduces computation costs and eliminates noise from the background. Nonetheless, the cropping operation can lead to information loss if the crop size is not chosen reasonably. 
We first estimate the crop size to represent the fovea area of human fixation following~\cite{rong2021human}.
Concretely, the fovea area has the radius of $r = tan 2^{\circ} \cdot d = 21~mm$. The eye-track screen has a horizontal direction a length of $530~mm$ and a resolution of $1920$ pixels. Hence, the radius contains $\frac{21}{530} * 1920 \approx 75$ pixels. We examined the hyper-parameters of the cropped size within the range of $[25, 50, 75, 100]$. We used the collected human gaze fixation in this comparison.
From Table \ref{tab:crop size}, we see that decreasing the crop size leads to information loss, resulting in a decline in performance, especially activity recognition where a constant decrease in performance can be observed ($0.63$ for $B=100$ and $0.55$ for $B=25$). However, increasing the crop size may introduce noise information from the background, potentially leading to a reduction in atomic action prediction, which requires more precise information input from the object detector. Interestingly, $75$ also corresponds to the human fovea area. We thus set $B=75$ for our model and in other experiments.

\begin{table}[t]
\caption{Results of using different crop sizes to represent the radius of the fovea area. The best results are marked in bold.}
\centering
\resizebox{.75\linewidth}{!}{
\begin{tabular}{c|c|c|c|c}
\toprule[1pt]
 Crop Size    & 25 & 50 & 75 & 100 \\ \hline
Recog. Acc. $\uparrow$ & 0.55    &  0.59  &  0.61 &   \textbf{0.63}     \\ \hline
Pred. IoU $\uparrow$ &  0.33  &  0.34  & \textbf{0.35}  &  0.33  \\\hline
Pred. Leven. $\downarrow$ & 0.58 & 0.56 & \textbf{0.51} & 0.54 \\ 

\bottomrule[1pt]
\end{tabular}
}
\label{tab:crop size}
\end{table}

\begin{figure}[ht]
    \centering
    \includegraphics[width=.7\linewidth]{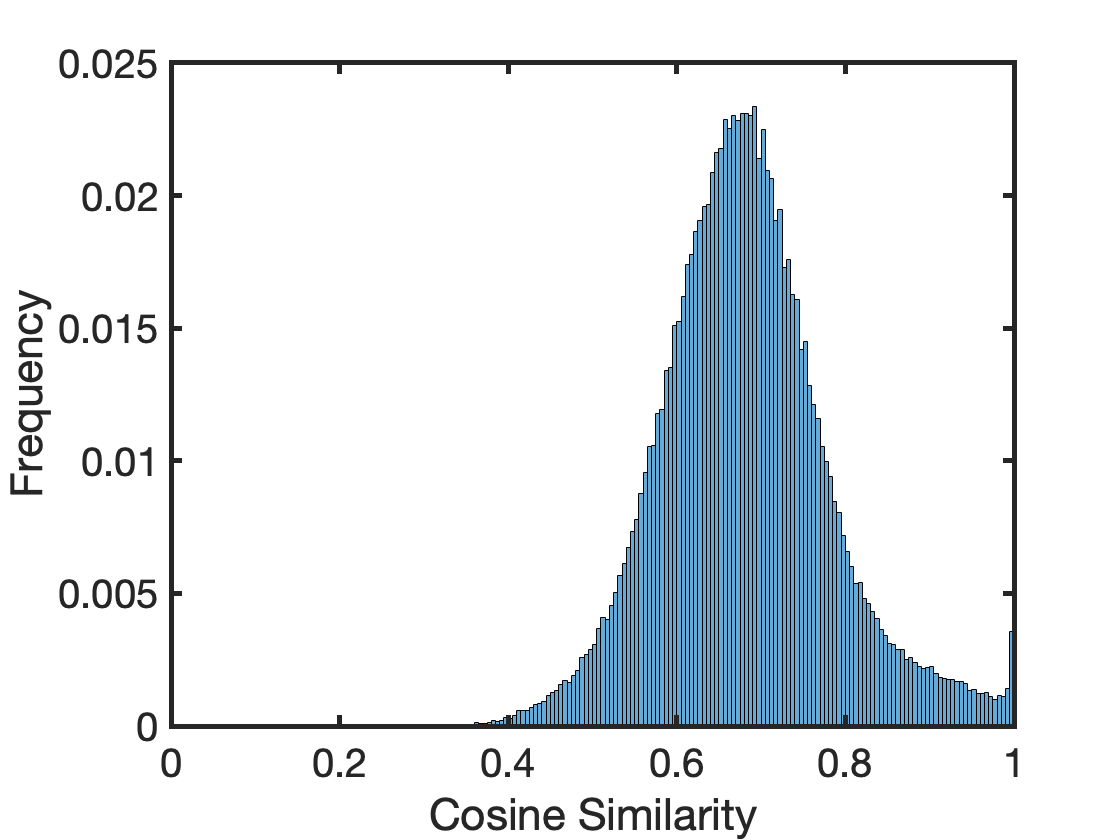}
    \caption{Histogram of cosine similarity values between two visual embeddings.}
    \Description{Histogram of cosine similarity values between two visual embeddings.}
    \label{app-fig:cosinehist}
\end{figure}

The threshold $\rho$ in Algorithm \ref{alg:video2graph} decides the similarity for merging visual embeddings into one node. 
To set the $\rho$, we calculate the cosine similarity between the visual embeddings of a frame (fovea area) to other embeddings inside the same video and plot the distribution of similarity values in Figure \ref{app-fig:cosinehist}. Based on the distribution, we decide to use $0.9$ as the threshold, since it is able to distinguish most of the frames but also does not create too many nodes for a graph.

\section{Implementation Details}
\label{sec:app-implementation}

\subsection{Activity recognition and prediction.} 
We utilized the trained model \texttt{CLIP}~\cite{radford2021learning} as our visual encoder and \texttt{Word2Vec}~\cite{mikolov2013efficient} as the semantic encoder. Hence, we obtained the node embedding $v_i\in \mathbb{R}^{512}$ and the edge embeddings $e_i\in \mathbb{R}^{600}$. A three-ECC layer GNN with the hidden dimension of $128$ was used to process the graph $\mathcal{G}$ to get the visual-semantic embeddings $v_{\mathcal{G}} \in \mathbb{R}^{384}$, which was fed into a linear layer for activity recognition, and an LSTM layer with the hidden dimension of $384$ followed by another linear layer for the atomic action prediction. The network was trained using the Adam optimizer for $300$ epochs. The initial learning rate was set to $1e^{-3}$.

\subsection{Object detector}
\label{sec:app-obj}
We generate new frames for training an object detector. We have $38$ objects (interacted by the agent in the video dataset) and we provide around $1000$ training samples to train the object detector for each object class. Concretely, we obtain the semantic maps for new frames provided by the VirtualHome platform, where each pixel is labeled with an object class. For one object class, we randomly sample a pixel with the label of that class. Then, we crop the corresponding RGB frame centered at that pixel using a bounding box size of $150$ pixels. The object detector consists of a visual feature encoder and a classifier. For feature encoding, we deploy the  pretrained CLIP model\cite{radford2021learning}. In the classifier, there are two linear layers and a dropout layer with a dropout rate of $0.8$. The input dimension of the linear layer is set to 512 with a hidden dimension of 128. We train our object detector with the cross-entropy loss using the Adam optimizer with the learning rate set to $1e{-4}$. The training loss converges after $300$ epochs. We test the object detector on around $200$ samples for each class and it achieves the accuracy of $0.72$ on the whole test set.

\section{More Experimental Results}
\label{sec:app-more results}

\subsection{More Ablation study}
We conduct another ablation study to verify the advantages of utilizing semantic components in graph modeling to enhance activity understanding. 

Compared to the previous works~\cite{nagarajan2020ego,hussein2019videograph} using graphs for activity recognition, we propose to fuse the object features into the graph modeling. The object features come from the fixated object as introduced in Section \ref{sec:method-gvsf}. 
To show the advantage of our proposed visual-object fusion, we first compare it to the baseline where solely visual features, i.e. node embeddings, are used, similar to the settings in \cite{nagarajan2020ego}. When comparing our results with the "Visual" only baseline in Table \ref{tab:semantic emb}, our model outperforms in both activity recognition and atomic action prediction tasks, e.g., decreasing the activity classification score by $7\%$ and increasing the Levenshtein distance in action prediction by $0.12$. 
This indicates that object information adds relevant class-specific information. We also evaluate a "random object" baseline, in which random object information is assigned to edges in the graph. This also results in performance degradation, especially in predicting action sequences such as the Levenshtein distance, highlighting the importance of incorporating relevant object information for accurate action prediction. Furthermore, this comparison reveals the precision of our object detector, which benefits the challenging action prediction task.

\begin{table}[ht]
\caption{Influence of object features (edge attributes) in activity recognition and the atomic action prediction tasks.}
\resizebox{0.95\linewidth}{!}{
\begin{tabular}{c|c|c|c}
\toprule[1pt]
    & Recog. Acc. $\uparrow$ & Pred. IoU $\uparrow$ & Pred. Leven.$\downarrow$  \\\hline
Visual  &  0.54  &   0.26   &    0.63   \\ \hline
Visual + Random Object &  0.59  &  0.31     &   0.60  \\ \hline
Visual + Object (Ours) & \textbf{0.61} & \textbf{0.35} &  \textbf{0.51} \\
\bottomrule[1pt]
\end{tabular}
}
\label{tab:semantic emb}
\end{table}

\subsection{More Quantitative Results} \label{app-more_quantitative}
We run the same evaluation on our methods as well as the CNN-based method using $50\%$ and $90\%$ of the video as the input in Table \ref{app-tab:50} and \ref{app-tab:90}, respectively. We observe that using longer videos as the input eases the task for models. Our model incorporated with human gaze fixation achieves the best performance, and our model outperforms the baseline method, highlighting the advantage of our proposed framework.
Table \ref{app-tab:comparisonmeanstd} lists the results of our model compared with the baseline model over three runs. The results are shown in (mean $\pm$ std).
\begin{table}[ht]
\caption{Results using $50\%$ of the video as input}
\resizebox{.95\linewidth}{!}{
\begin{tabular}{c|c|c|c}
\toprule[1pt]
    & Recog. Acc. $\uparrow$ & Pred. IoU $\uparrow$ & Pred. Leven. $\downarrow$ \\\hline
IGA \cite{min2021integrating} & 0.23   &  0.11 &    0.77     \\ \hline
Ours  & \textbf{0.41}  & \textbf{0.35} & \textbf{0.59}  \\ 
\bottomrule[1pt]
\end{tabular}
}
\label{app-tab:50}
\end{table}

\begin{table}[ht]
\caption{Results using $90\%$ of the video as input}
\centering
\resizebox{.95\linewidth}{!}{
\begin{tabular}{c|c|c|c}
\toprule[1pt]
    & Recog. Acc. $\uparrow$ & Pred. IoU $\uparrow$ & Pred. Leven. $\downarrow$ \\\hline
IGA \cite{min2021integrating} & 0.33   &  0.24 &   0.68     \\ \hline
Ours  & \textbf{0.63}  & \textbf{0.47} & \textbf{0.49} \\ 
\bottomrule[1pt]
\end{tabular}
}
\label{app-tab:90}
\end{table}

\begin{table}
\caption{Results of different models over three runs. (Mean $\pm$ Std) is shown.}
\centering
\resizebox{.95\linewidth}{!}{
\begin{tabular}{c|c|c|c|c}
\toprule[1pt]
    & Recog. Acc. $\uparrow$ & Pred. IoU $\uparrow$ & Pred. Leven. $\downarrow$ &  Success Rate $\uparrow$\\\hline
Human & \textbf{0.94} & - & - & - \\\hline
IGA \cite{min2021integrating} &  0.27$\pm$0.004  & 0.21$\pm$0.001 &   0.75 $\pm$0.002 &  0.14$\pm$0.002    \\ \hline
Ours  &0.60$\pm$0.010& \textbf{0.36}$\pm$0.015 & \textbf{0.52}$\pm$0.007 & \textbf{0.26}$\pm$0.014  \\ 
\bottomrule[1pt]
\end{tabular}
}
\label{app-tab:comparisonmeanstd}
\end{table}

\subsection{More Qualitative Results on Action Anticipation}\label{app-more-qualitative}
\begin{figure}[ht]
    \centering
    \includegraphics[width=0.6\linewidth]{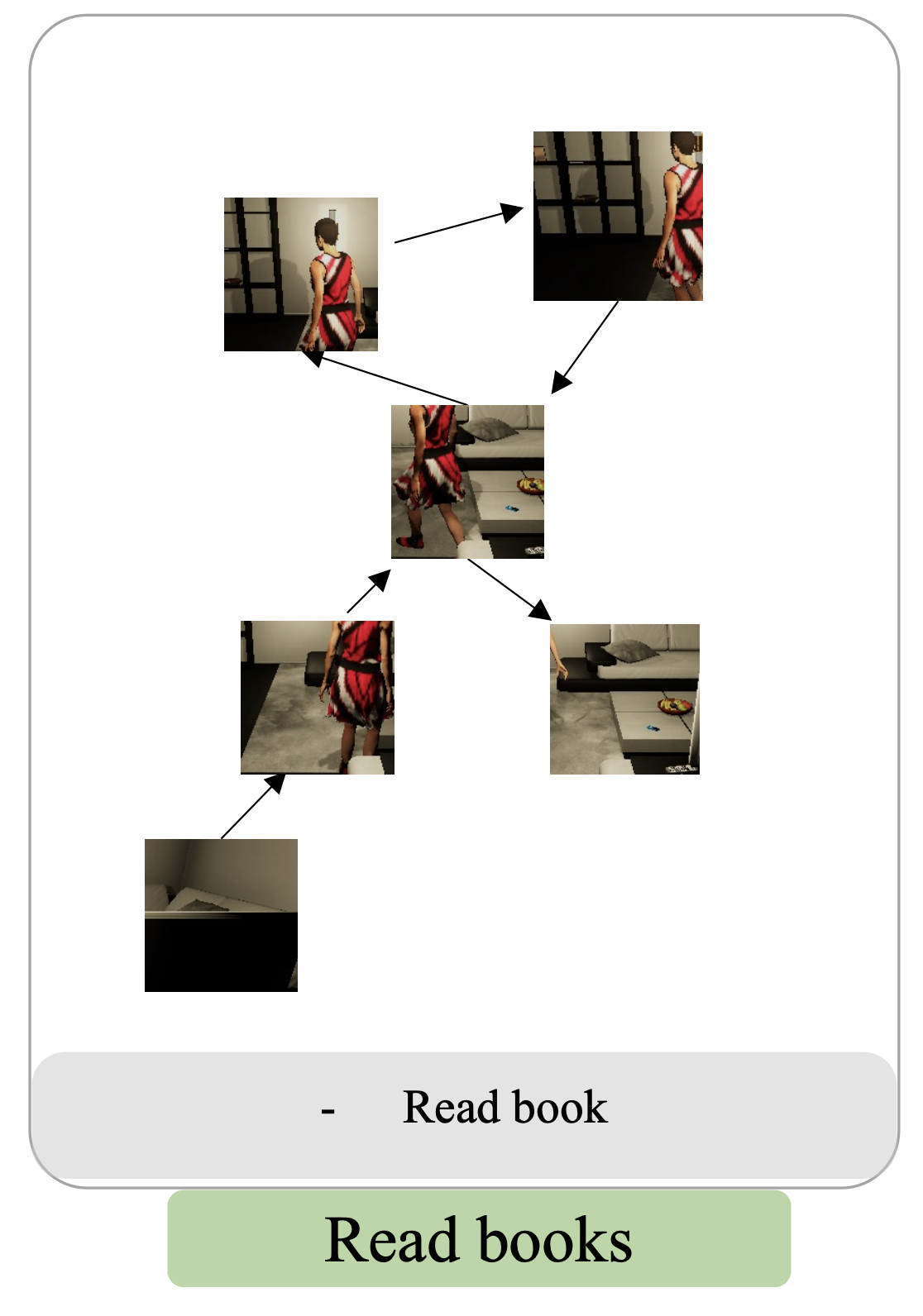}
    \caption{Qualitative result for the activity "Read books".}
    \Description{Qualitative result for the activity "Read books".}
    \label{app-fig:graph_example2}
\end{figure}

\begin{figure}
    \centering
    \includegraphics[width=0.68\linewidth]{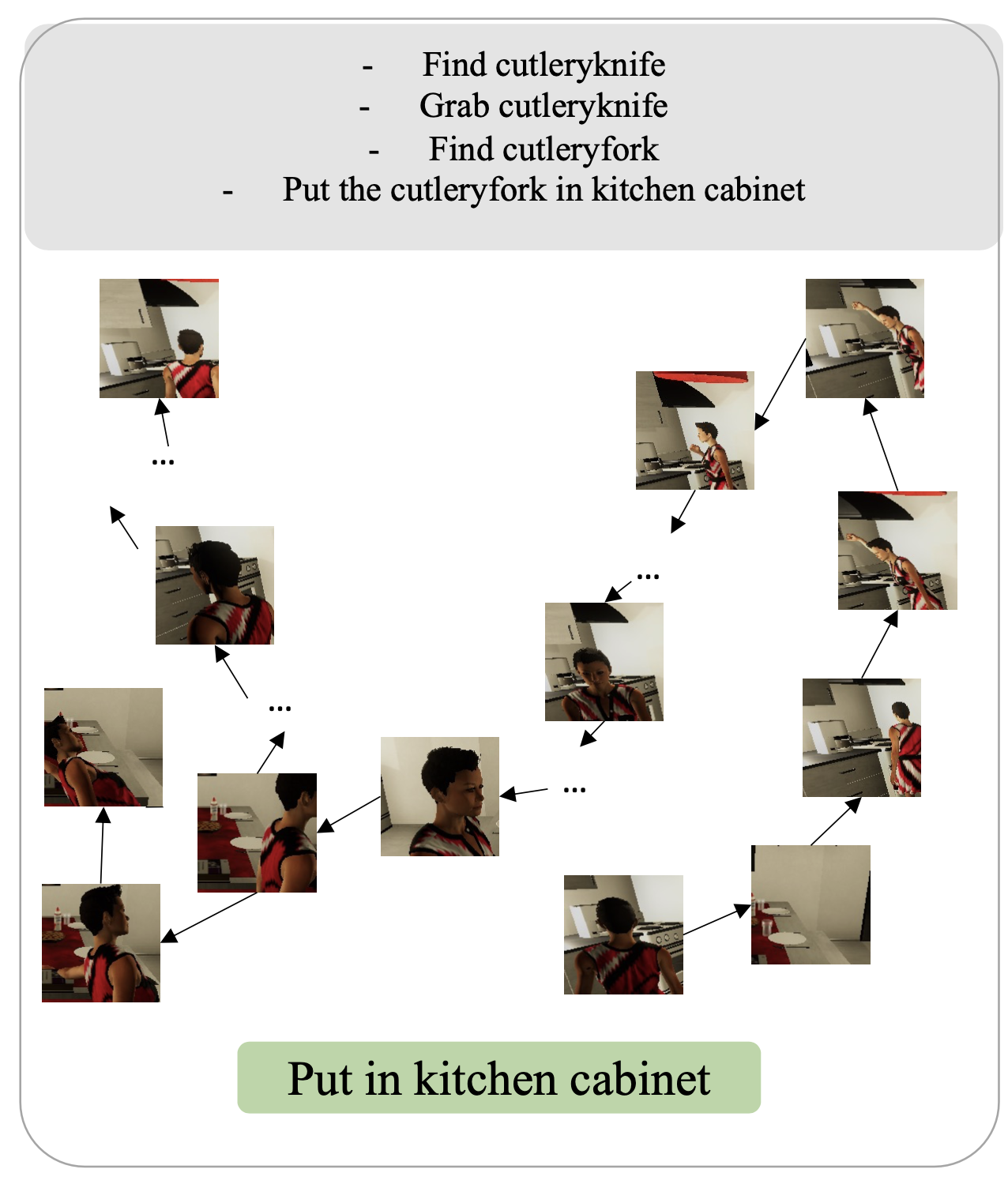}
    \caption{Qualitative result for the activity "Put the cutlery in the kitchen cabinet".}
    \Description{Qualitative result for the activity "Put the cutlery in the kitchen cabinet".}
    \label{app-fig:graph_example3}
\end{figure}

More qualitative results of our whole framework are shown in Figure \ref{app-fig:graph_example2} for the activity "Read books" and Figure \ref{app-fig:graph_example3} for the activity "Put the cutlery in the kitchen cabinet".

\end{document}